\documentclass[a4paper, 10pt, conference]{ieeeconf}      

\IEEEoverridecommandlockouts                              
\usepackage{graphics} 
\usepackage{epsfig} 
\usepackage{mathptmx} 
\usepackage{times} 
\usepackage{amsmath} 
\usepackage{amssymb}  
\usepackage{xcolor}
\usepackage{cite}
\usepackage{verbatim}

\title{\LARGE \bf
An origami crawling robot driven by a folded self-sustained oscillator}

\author{Wenzhong Yan$^{1,\dag}$ and Ankur Mehta$^{2}$
\thanks{$^{1}$Wenzhong Yan is with Department of Mechanical and Aerospace Engineering,
        University of California, Los Angeles, CA 90095, USA. $^{2}$Ankur Mehta is with the Department of Electrical and Computer Engineering, University of California, Los Angeles, CA 90095, USA. }%
\thanks{$^{\dag}$ Corresponding author, {\tt\small wzyan24@g.ucla.edu}}%
}

\begin{document}

\maketitle
\thispagestyle{empty}
\pagestyle{empty}

\begin{abstract}
Locomotive robots that do not rely on electronics and/or electromagnetic components will open up new perspectives and applications for robotics. However, these robots usually involve complicated and tedious fabrication processes, limiting their applications.
Here, we develop an easy-to-fabricate crawling robot by embedding simple control and actuation into origami-inspired mechanisms through folding, eliminating the need for discrete electronics and transducers. 
Our crawling robot locomotes through directional friction propelled by an onboard origami self-sustained oscillator, which generates periodic actuation from a single source of constant power. The crawling robot is lightweight ($\sim$3.8 gram), ultra low-cost ($\sim$US \$1), nonmagnetic, and electronic-free; it may enable practical applications in extreme environments, e.g., large radiation or magnetic fields. The robot can be fabricated through a monolithic origami-inspired folding-based method with universal materials, i.e., sheet materials and conductive threads. This rapid design and fabrication approach enables the programmable assembly of various mechanisms within this manufacturing paradigm, laying the foundation for autonomous, untethered robots without requiring electronics. 
\end{abstract}
\section{INTRODUCTION}


Robots capable of locomotion on ground play an important role on expanding the exploration space for humans; these robots provide the access to extreme environments or dangerous terrain that are unreachable to humans. These mobile robots have attracted much interest over the past decades. However, they usually require complicated fabrication process and incorporate electronics and/or electromagnetic components, e.g., inductors and electromagnetic motors, for control and actuation to achieve locomotion. These requirements could limit the application of mobile robots: i) the complex design and fabrication process increases the cost and difficulty of the creation of robots, which could in turn hinder the advance and development of robotics; ii) the dependency on electronics and/or electromagnetic components may restrict the applications in extreme areas, e.g., environments with high magnetic or radiation. Therefore, we desire mobile robots that can be created in a rapid prototyping manner and without requiring electronics and/or electromagnetic components, which could further extending the design space and application scenarios of robotics. 

\begin{figure}[t]
    \centering
    \includegraphics[trim= 1.9in 22.3cm 1.7in 0cm, clip=true, width=3.3in]{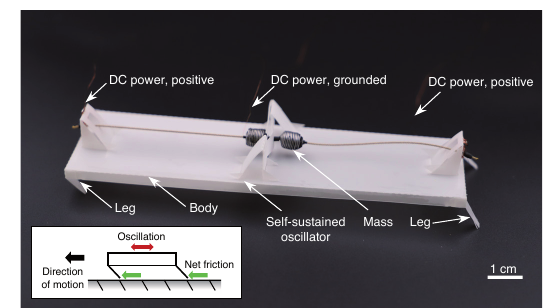}
    \caption{A proposed origami-inspired crawling robot driven by a folded self-sustained oscillator ($\sim$3.8 g, $138 \times 38 \times 22$ mm). This robot is mainly composed of an origami body, four angled legs and one self-sustained oscillator. The back-and-forth movement of the oscillator drives the robot to move forward through directional friction. A nonmagnetic 2-gram mass made of tin is attached on the bistable beam to facilitate locomotion. A power supply with constant current is provided through tethers.}
    \label{fig:robot}
\end{figure}

There have been increasing efforts spent on creating electronics-free locomotive autonomous robots. Octobot, a completely soft autonomous robot, uses microfluidic logic as an on-board controller, eliminating the need of electronics, enabling untethered operation autonomously\cite{wehner2016integrated}. Macro-scale electronics-free robots are designed and created by harnessing a soft, elastomeric monostable valve to regulate air flow for controlled actuation, allowing peristaltic and rolling locomotion of soft robots \cite{rothemund2018soft,preston2019soft}. Afterwards, soft-legged robots with sophisticated gait control \cite{drotman2021electronics} are developed based on the pneumatic circuits. 
A new class of soft robots based on dielectric elastomer actuators (DEAs) are able to generate crawling locomotion by only using a constant high-voltage power source without electronics\cite{henke2017soft}. RoBeetles, insect-scale autonomous crawling robots, incorporate a mechanical control mechanism with catalytic artificial micromuscle\cite{yang202088}. This class of electronics-free locomotive robots often relies on oscillation, which arises from special configurations with relatively constant energy inputs, including pneumatic sources\cite{xu2020electronics}, humility discrepancies\cite{chen2015scaling}, light\cite{gelebart2017making}, temperature gradients\cite{kotikian2019untethered}, etc. However, these robots often involve complicated and tedious fabrication processes. For example, molding and sealing are always very challenging issues for pneumatic robots. Therefore, simpler and more accessible design and fabrication methods to achieve locomotion on ground are desired.  

Folding-based manufacturing as a top-down, parallel transformation approach is a very promising method to simplify and accelerate the design and fabrication of autonomous locomotive robots without the need of electronics and/or electromagnetic components. This method is inspired by the ancient art of origami, which takes advantage of available 2D planar materials and manufacturing machines to construct 3D structures and devices
\cite{onal2012origami,martinez2012elastomeric,min2008geometrical,kim2018origami}. However, current origami robots have mainly focused on mechanical mechanisms and rely on off-the-shelf electronics and/or electromagnetic actuators, like micro-controllers and DC motors, for locomotion\cite{rus2018design}. 
There are several attempts to develop control hardware\cite{miyashita2014self,treml_origami_2018} by using the origami-inspired method, however, creating integrated autonomous robots capable of locomotion on ground without the need of electronics and/or electromagnetic components is still challenging.

Towards our ultimate goal of realizing origami autonomous robots without the need of electronics, here, we propose an origami crawling robot (see Fig.\ref{fig:robot}) by embedding simple control (oscillation) and actuation directly into origami-inspired mechanisms, eliminating all discrete electronics and actuators. This class of origami robots, only requiring universal raw materials and easy to be fabricated, will further advance the inexpensive and rapid prototyping of robots. Meanwhile, the electronic-free nature of these robots would be applicable for applications where traditional semiconductor-based components may not survive. In addition, the nonmagnetic property could potentially complement the motor-mediated robots in high magnetic fields.


The crawling robot (Fig.\ref{fig:robot}) builds upon our preliminary work on the design of origami self-sustained oscillator\cite{yan2018towards,yan2021cut} to achieve a much more challenging and meaningful task of directional crawling on ground; we present here the integration of this actuator into a complete crawling robot. This robot can be later integrated with on-board power to create untethered robots to realize the full potential of origami-inspired robotics. Specifically, the contributions of this paper include:
\begin{itemize}
    \item a locomotion mechanism through the combination of directional friction and transient impact-induced oscillation,
    \item a method that enables building functional robotic systems purely through cut-and-fold, only requiring universal materials, and
    \item an origami crawling robot fabricated out of the proposed design and verified by experiments.
\end{itemize}


The remainder of the paper is organized as follows: in Section~\ref{origamiOscillator}, we briefly introduce the origami-inspired self-sustained oscillator; in Section~\ref{theRobot}, we introduce the design, mechanism, simplified model and performance of the crawling robot; and the conclusion is presented in Section~\ref{Conclusion}.

\section{Origami-inspired self-sustained oscillator}
\label{origamiOscillator}
\begin{figure}[t]
    \centering
    \includegraphics[trim= 1.9in 12.6cm 1.7in 0cm, clip=true, width=3.1in]{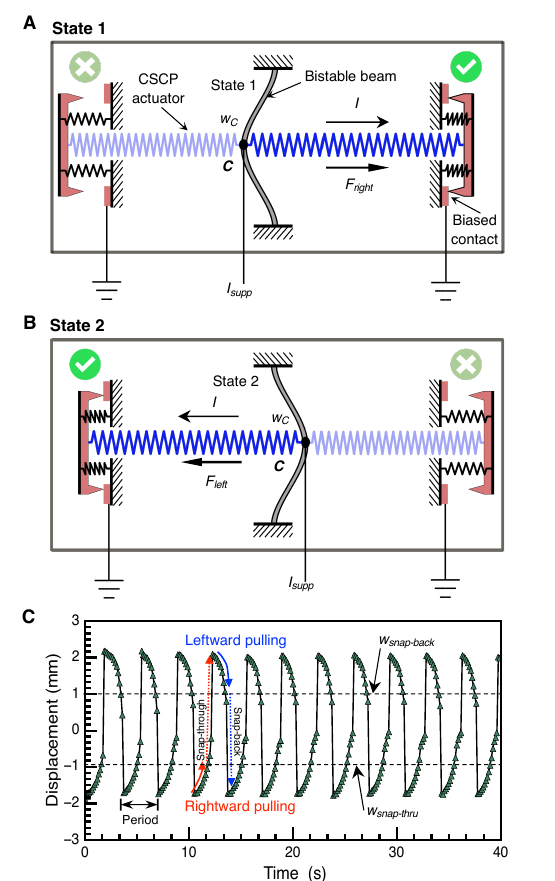}
    \caption{Operation of the  origami self-sustained oscillator. The schematic of the oscillator is shown in (A) and (B). (C) The oscillator generates an alternating output motion of the bistable beam when a DC electric power supply ($I\textsubscript{supp}$ = 0.62 A) is applied. The oscillation period is about 3.4 s.}
    \label{fig:oscillator}
\end{figure}
\begin{figure}[t]
    \centering
    \includegraphics[trim= 1.9in 20.8cm 1.7in 0cm, clip=true, width=2.8in]{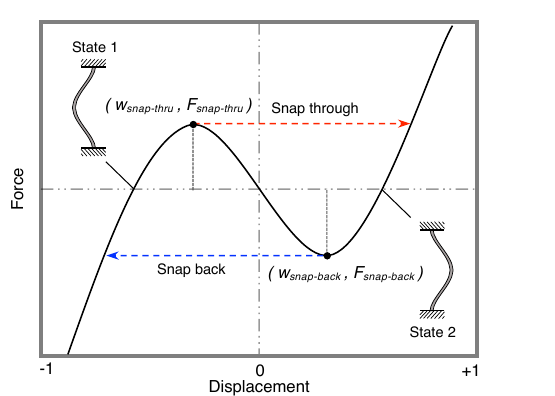}
    \caption{The bifurcation diagram of the bistable beam. The beam has two stable equilibrium states and can transit between each other.}
    \label{fig:bistableMechanism}
\end{figure}

The origami oscillator can generate periodic motion with a constant DC power without requiring control (Fig.\ref{fig:oscillator}). It mainly consists of a bistable buckled beam, two conductive super-coiled polymer (CSCP) actuators\cite{yip_control_2017}, and two biased contacts. One end of each actuator is attached on the center point C of the bistable beam while the other is attached on the biased contact. When the oscillator is at state 1 (Fig.\ref{fig:oscillator}A), the right biased contact is closed and the actuator is powered, generating a pulling force to drive the bistable beam snap-through (see Fig.\ref{fig:bistableMechanism}). Once the bistable beam snaps through, the left actuator is powered while the right is disconnected and starts to cool down (Fig.\ref{fig:oscillator}B). After reaching the critical point, the bistable beam snaps back to state 1 and thus the oscillator returns to its initial state and begins the next cycle of operation. As time evolves, the bistable beam of the oscillator will snap back and forth horizontally between its two stable states, i.e. state 1 and state 2 in Fig. \ref{fig:oscillator}, featuring a transient impact-induced reciprocating actuation. A detailed mechanism design and analysis can be found in \cite{yan2018towards,yan2021cut}.

Figure \ref{fig:oscillator}C demonstrates a typical motion of the oscillator under a 0.62 A constant power supply. The displacement curve represents the movement of the center point C on the bistable beam; the data was extracted from videos by using an analysis software--Tracker. It features an oscillation period of about 3.4 s with the amplitude of $\sim$ 1 mm.

\section{Design and fabrication of the crawling robot}
\label{theRobot}
\subsection{Design and Mechanism}
\begin{figure}[t]
    \centering
    \includegraphics[trim= 1.9in 18.7cm 1.7in 0cm, clip=true, width=3.3in]{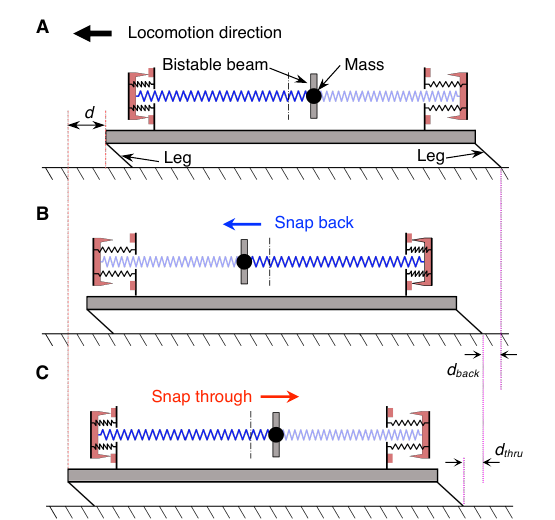}
    \caption{Oscillation-driven crawling mechanism of the robot. (A) Initially, the mass stays at the right equilibrium state, i.e., state 2. Once the bistable beam is pulled to reach the critical point, the mass snap back (leftward in the figure), its kinetic energy drive the robot move leftward by a distance, $d\textsubscript{back}$ (B). (C) Similarly, when the mass snap through (rightward in the figure), the robot also moves leftward by a distance, $d\textsubscript{thru}$ due to the combination of the impact-induced actuation and special polyvinyl chloride surface of the ground. After one oscillation, the mass backs to its original position and thus the robot moves forwards by a net distance d.}
    \label{fig:locomotionMechanism}
\end{figure}

The origami crawling robot, as shown in Fig.\ref{fig:robot}, mainly consists of an origami body, four angled legs, and one oscillator (with its attached mass). These angled legs (see Fig. \ref{fig:locomotionMechanism}A) are adopted due to two reasons: First, they are used to reduce the contact area of the robot body with the ground to reduce friction. Second, each leg contacts the ground with an angle of ~$45^\circ$ because this angle is suitable for generating directional friction on a broad range of surfaces\cite{ji2019autonomous}. The oscillator is integrated onto the robot, siting on the top surface of the body (Fig.\ref{fig:locomotionMechanism}A). We further attached a two-gram mass to the bistable beam. This mass is added to improve the transmission efficiency from the potential energy of the bistable beam to the kinetic energy of the robot, increasing its locomotion speed. However, the increment of the weight of the robot increases the friction. Therefore, the weight of the mass needs to be designed carefully to optimize the locomotion speed of the robot. In this paper, we use a two-gram mass, which is empirically determined for now.

The simplified oscillation-driven crawling locomotion mechanism of this robot can be described in two consecutive steps. In the first step, the bistable beam (with the mass) snaps leftwards (Fig.\ref{fig:locomotionMechanism}B). The kinetic energy of the mass is delivered to the robot, driving it leftward for a certain distance, \textit{d\textsubscript{back}}. For the second step as shown in Fig.\ref{fig:locomotionMechanism}C, the robot moves further forward by another distance, \textit{d\textsubscript{thru}}, due to the reaction from the bistable beam snap-through to the right. The backward motions in each step are suppressed due to the asymmetric friction against the surface condition of the selected terrain---a polyvinyl chloride cutting mat (10671, Dahle Vantage) --- and the unique transient impact-induced actuation from the snap-through. Thus, the robot can totally move forward by a distance, $d (=\textit{d\textsubscript{back}} + \textit{d\textsubscript{thru}})$, in one oscillation cycle (see Fig.\ref{fig:locomotionMechanism}C). As time evolves, the robot can continuously crawl through directional friction driven by the oscillation-induced actuation.

\subsection{Simplified Analytical Model}
\label{model}
To better understand the mobility of the crawling robot, we build a simplified analytical model of its motion. We consider the robot has a lumped mass, $M+m$, while the $m$ is contributed by the attached mass on the bistable beam. The legs of the robot are rigid, joined to the body with a infinite rotational spring of stiffness, while we assume that their legs are in frictional contact with the substrate. The friction is simplified as Coulomb friction model. Thus, the friction force when sliding is $F_f =\mu N$, where $N$ is the normal force against the substrate. 

As described above, the robot stays still during the pulling process. Once the bistable beam with its attached mass reaches the critical snapping point, the stored bending energy, $E$, is released instantly and converted to kinetic energy of the beam and the robot with some loss (e.g., damping). The bending energy, $E$, is exclusively determined by the geometry and material properties of the bistable beam\cite{yan2019analytical}. To simplify this complicated process, we assume the bending energy, $E$, could be converted into kinetic energy with the coefficient of efficiency, $\eta$. We also assume the attached mass reaches its maximum speed right after the snap-through/back and instantly collides and merge with the body of the robot; this assumption will overestimate the converted kinetic energy. Thus, the initial velocity of the robot can be approximated as:
\begin{equation}
   \nu_{init} = \sqrt{\frac{2\eta E}{m}} \frac{m}{M+m}
\label{eq:velocity}
\end{equation}

With this initial velocity, the robot can glide on the frictional substrate with a distance:
\begin{equation}
   s = \frac{{\nu_{init}}^2}{2\mu g}
\label{eq:distance}
\end{equation}

Assuming the robot can move the same distance during both the snap-through and snap-back of the bistable beam, we can have the average velocity of the robot:
\begin{equation}
   \bar{\nu} = \frac{2s}{T_{osc}}\\
   = \frac{2 \eta E }{\mu g (\frac{M}{m}+1)^2 T_{osc}}
\label{eq:avarageSpeed}
\end{equation}

It is worth noting that the oscillation period, $T_{osc}$, is mainly determined by the geometry and material properties of the bistable beam, CSCP actuators, and the supply power\cite{yan2019rapid}. This simplified model overestimates the average velocity of the robot due to the assumptions; it can qualitatively predict the speed of robot. As the model suggested, to increase the speed, we can enlarge the bending energy $E$ and/or the weight of attached mass, or decrease the coefficient of friction against the substrate and/or oscillation period. Accurate models are necessary for quantitative prediction.

\subsection{Fabrication and Assembly}
The creation of the proposed origami crawling robot is a systematic method of incorporating different components to generate an integrated robotic system without requiring electronics. It mainly has two consecutive steps: (i) origami cut-and-fold and (ii) integration and assembly with actuators and necessary accessories\cite{yan2020towards}. 

\begin{figure}[t]
    \centering
    \includegraphics[trim= 1.9in 18cm 1.9in 0cm, clip=true, width=3.2in]{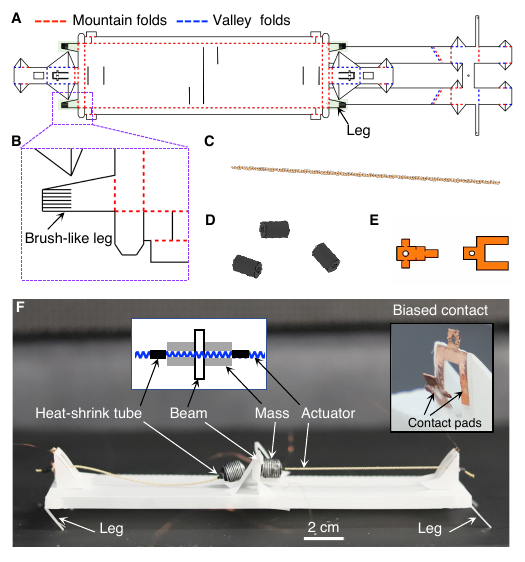}
    \caption{Design and assembly of the origami crawling robot. (A) 2D fabrication pattern of the origami frame of the robot. The patterns of the legs are in green shadow. (B) The zoom-in view of the brush-like angled legs. (C) CSCP actuators functioning as both conductor and actuator. (D) Heat-shrink tube to fix the position of actuators. (E) Copper contact pads forming the conductive interface of the biased contact. (F) The detailed structure of the crawling robot. Insert: side view of the biased contact.}
    \label{fig:fabrication}
\end{figure}

\subsubsection{Origami Cut-and-Fold}
We fabricated the body of the robot by patterning a thin polyester (PET) film with a Silhouette CAMEO. The 2D pattern of the crawling robot is shown in Fig. \ref{fig:fabrication}A. In this pattern, red dashed lines mean mountain folds and blue dashed lines represent valley folds. The angled legged are highlighted in green shadow. By following the folding pattern, the final 3D model of the origami framing was created. Specifically, the four legs are folded along the edges to form angled contact with the ground to generate directional friction. For now, we shaped the geometry and angle of the leg empirically; this may introduce fabrication errors. In the future, a systematic computational design method may result in greater control over the behavior of the robot through tuning configuration of the legs. The detailed features of the leg are presented in Fig.\ref{fig:fabrication}B. A brush-like leg design was adopted to increase its tip's flexibility; the increased differentiation of the coefficients of friction in each direction improves the resulting locomotion of the robot. In addition, the flexibility of the tip of the leg can improve the stability of the robot. However, it is challenging to model the contact angle when the leg tip becomes too soft; instead data-driven approaches may be called for.

\subsubsection{Integration and Assembly}
After the origami framing been built, two CSCP actuators, heat-shrink tubes, contact pads, and other accessories (Fig. \ref{fig:fabrication}C to E) are integrated into the system to complete the robot. To improve the transmission efficiency of the system, a mass (two tin balls with central holes, though any mass will suffice) was mounted around the CSCP actuators beside the bistable beam, using heat-shrink tubes to form mechanical stoppers, as shown in Fig. \ref{fig:fabrication}F.
This monolithic fabrication process allows the creation of robots built out of a single sheet of PET sheet in a single step, which enables easy and rapid prototyping.

\subsection{Locomotion and Analysis}
\begin{figure}[t]
    \centering
    \includegraphics[trim= 1.7in 17.3cm 1.7in 0cm, clip=true, width=3.3in]{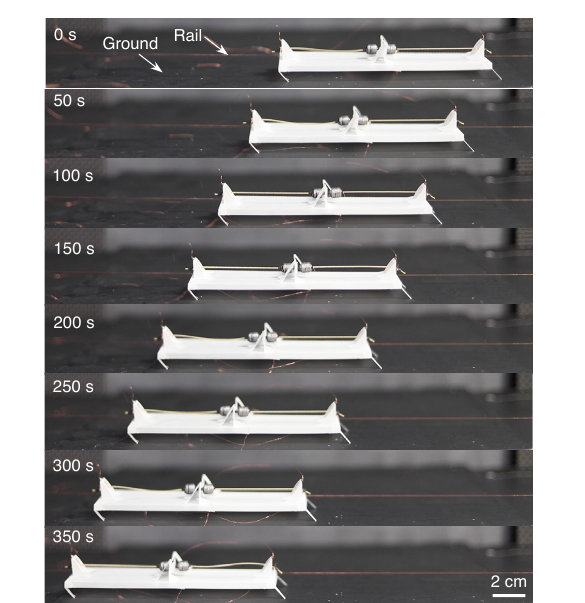}
    \caption{The oscillation-driven crawling locomotion of the robot is presented in a sequence of snapshots, from top to bottom. A thin copper wire is used to serve as a guiding rail for the robot. The robot only requires a constant power source without electronics or control.}
    \label{fig:crawlingSnapshot}
\end{figure}
The total material cost of the origami crawling robot made by our proposed approach is about US\$1. In addition, fabrication takes only about 1.5 hours for a practiced maker. Before the experiment, we used a thin copper wire to create a rail to guide the locomotion direction of the robot (see Fig. \ref{fig:crawlingSnapshot}), though the additional friction from the rail may reduce the locomotion speed. The robot was powered with a constant current by a laboratory power supply tethered through thin copper wires. In addition, a cooling system was equipped to control the frequency of the oscillator. The cooling system was placed at an approximately 150 mm distance from the tested devices; its parameters were set to about 110 cfm flow rate and $\sim 22^\circ$C.

\begin{figure}[t]
    \centering
    \includegraphics[trim= 2.3in 15.8cm 2.3in 0.5cm, clip=true, width=3.3in]{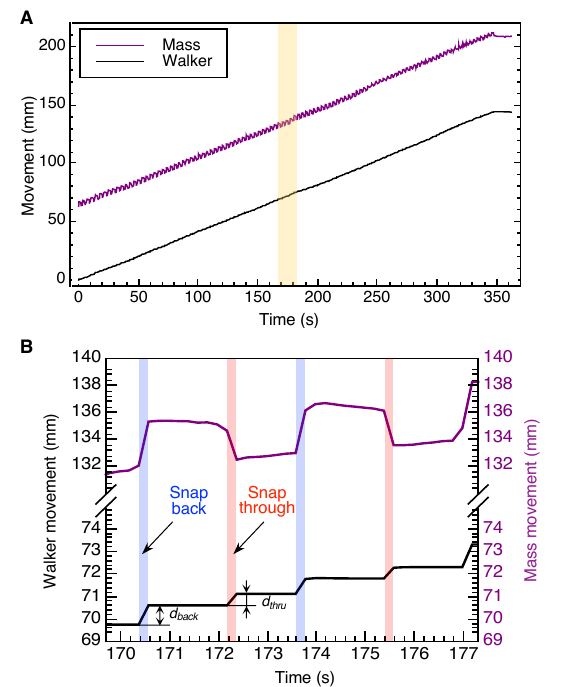}
    \caption{Time-resolved plots of the displacement of the crawling robot over time, extracted from the video. The robot is able to monotonically locomote 146.0 mm in 350.1 s, achieving an average speed of about 25.2 mm/min (about 0.18 body lengths/min). (B) Zoom-in movement curves from 170 s to 177 s of (A).}
    \label{fig:displacementCurve}
\end{figure}

The coefficients of friction in two directions are measured as 0.36 (forward) and 0.72 (backward). After a 10-second's rest, the robot was powered by a constant current supply. The mass started to snap back and forth to drive the robot move forward through directional friction, as shown in Fig. \ref{fig:crawlingSnapshot}. We traced both the trailing edge of the robot and the that of the mass and extracted visual information from the video to obtain their displacement curves as shown in Fig.\ref{fig:displacementCurve}A. The origami walker can monotonously march 146 mm in 350 s, achieving an average speed of about 0.42 mm/s, while the mass oscillatorily moved with the period of the mass about 3.3 s with the same average speed (Fig. \ref{fig:displacementCurve}B, and movie S1).

We further zoomed in the movement curves from 170 s to 177 s. The result validates that the back-and-forth motion of the oscillator is capable of realizing monotonous directional locomotion for origami robots (Fig.\ref{fig:displacementCurve}B). As mentioned above, the rightward snap-through motion of the mass could also drive the robot move leftward due to the special transient impact-induced actuation from the snap-through. However, this displacement, $d_{thru}$, is usually smaller than that caused by the leftward snap-back (Fig.\ref{fig:displacementCurve}B). Although it is out of the scope of this paper, this intriguing phenomenon may inspire new locomotion strategies for small-scale robots and open a new diagram of robotic research. Also, we notice that the crawling of the robot only happened when the snap-through (or snap-back) motion occurred, otherwise the robot stayed static. This phenomenon, in turn, validates our theory of the transient impact-induced locomotion mechanism of the walker as described in Section \ref{model}. 

\begin{figure}[t]
    \centering
    \includegraphics[trim= 1.5in 19.2cm 1.7in 0cm, clip=true, width=3.3in]{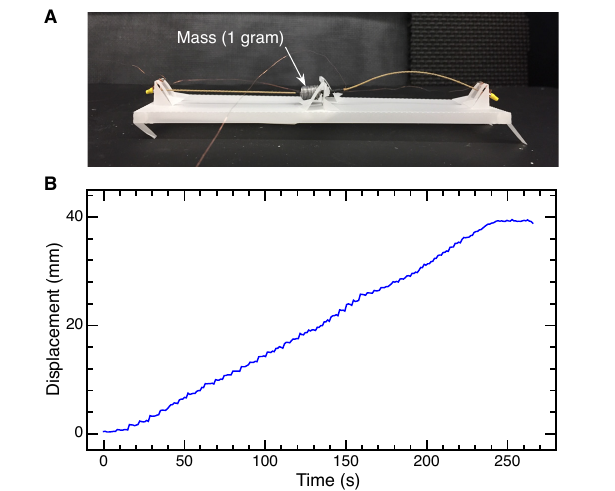}
    \caption{The origami crawling robot with one-gram attached mass on the bistable beam. (A) The detailed structure of the robot. (B) The displacement-time curve, extracted from the video.}
    \label{fig:robotOneGram}
\end{figure}

By reducing the mass to 1 gram, we ran the same test (see Fig.\ref{fig:robotOneGram}A). The result shows that the robot behaved similarly; it could move 39 mm in 4 minutes, achieving an average speed of about 0.16 mm/s (see Fig.\ref{fig:robotOneGram}B). This result qualitatively validates our simplified analytical model of the motion of the robot as described in Eq.\ref{eq:avarageSpeed}. Quantitatively, the analytical model suggests a 2.2 times of speed reduction while the actual speed of the robot decreased by 2.6 times, which indicates the viability of our model although further improvement of the model is desirable for accurate prediction.

The speed of the current walker is relatively slow in our experiment. Although not within the scope of this paper, we can speed up the robot with several methods as Eq.\ref{eq:avarageSpeed} suggests. The first one is to increase the oscillation frequency. We can also increase the amount of the released energy, $E$, of the bistable beam in each snap-through. For example, we can increase the width of the bistable beam or instead use stiffer membranes. In addition, we can investigate the effect of different surfaces on the speed of the robot. These options can be further investigated when the target is to optimize the locomotion speed of such robots. In this paper, we only chose the design that is easy to fabricate and control.

The crawling robot was powered by a 0.62A constant current supply; the average resistance of each actuator is about 3.8 ohm. The estimated power consumption is about 1.46 W. The efficiency of the current robot is low due to the thermal actuation. However, eliminating control and other accessories for achieving such locomotion can substantially reduce the weight and system complexity. 

\section{CONCLUSIONS}
\label{Conclusion}
We demonstrated an integrated, functional robot by embedding control and actuation into origami-inspired mechanisms, eliminating electronics for achieving meaningful tasks, i.e., locomotion on ground. The crawling robot can move through directional friction propelled by an on-board origami oscillator, which generates oscillation from a single source of constant power. This robot represents a new class of robots that expand the design space of origami folding-based manufacturing. The inexpensive and rapid prototyping nature, that only requires universal materials, improves the accessibility of robot creation. In addition, the non-electronic and amagnetic characteristic of this class of robots could enable applications in challenging environments, e.g., large magnetic or radiation fields, that would otherwise disturb typical metallic or semiconductor electronic components. In addition, our robot is environmentally friendly due to the electronics-free nature. 

The transient impact-induced driving mechanism of the robot may lead to new research on the locomotion of legged robots. This simple but effective locomotion strategy can greatly benefit small-scale robots subject to extremely limited payload capability and challenging fabrication. Before that, study on improving the energy efficiency of oscillation and thus driving should be conducted to realize practical applications. 

Towards our ultimate goal of creating untethered autonomous origami robots, we plan to integrate an on-board power source (e.g. paper battery \cite{hu2009highly}). The requirement for relatively high currents (but low voltages) for thermal actuation, however, would be a main challenge that limits the choice of its power supply; this problem is an established challenge in robotics in general, with a number of potential solutions actively under investigation. We believe that our origami crawling robots will be the basis of creating inexpensive, highly integrated, and autonomous robots.

\addtolength{\textheight}{0cm}





\section*{ACKNOWLEDGMENT}
This work is partially supported by the National Science Foundation under grant \#1752575. 


\bibliographystyle{IEEEtran}
\bibliography{Reference}

\end{document}